\documentclass[sigconf]{acmart}
\renewcommand\footnotetextcopyrightpermission[1]{} 
\usepackage{booktabs} 
\usepackage{bm}
\usepackage{algorithm,algorithmic}
\usepackage{mathtools}
\usepackage{subcaption}
\setcopyright{rightsretained}

\acmDOI{10.1145/nnnnnnn.nnnnnnn}

\acmISBN{978-x-xxxx-xxxx-x/YY/MM}

\acmConference[GECCO '18]{the Genetic and Evolutionary Computation
Conference 2018}{July 15--19, 2018}{Kyoto, Japan}
\acmYear{2018}
\copyrightyear{2018}

\acmArticle{4}
\acmPrice{15.00}

\editor{Jennifer B. Sartor}
\editor{Theo D'Hondt}
\editor{Wolfgang De Meuter}

\begin{document}
	\title{From Nodes to Networks: Evolving Recurrent Neural Networks}
	\author{Aditya Rawal}
	\affiliation{%
	    \institution{Univerisity of Texas Austin, Sentient Technologies, Inc}
	}
	\email{aditya@cs.utexas.edu}
	
	\author{Risto Miikkulainen}
	\affiliation{%
		\institution{Univerisity of Texas Austin, Sentient Technologies, Inc}
    }
	\email{risto.miikkulainen@sentient.ai}
	
	\begin{abstract}
		Gated recurrent networks such as those composed of Long Short-Term
		Memory (LSTM) nodes have recently been used to improve state of the
		art in many sequential processing tasks such as speech recognition and
		machine translation. However, the basic structure of the LSTM node is
		essentially the same as when it was first conceived 25 years ago.
		Recently, evolutionary and reinforcement learning mechanisms have been
		employed to create new variations of this structure. This paper
		proposes a new method, evolution of a tree-based encoding of the gated
		memory nodes, and shows that it makes it possible to explore new
		variations more effectively than other methods. The method discovers
		nodes with multiple recurrent paths and multiple memory cells, which
		lead to significant improvement in the standard language modeling
		benchmark task.  The paper also shows how the search process can be
		speeded up by training an LSTM network to estimate performance of
		candidate structures, and by encouraging exploration of novel
		solutions.  Thus, evolutionary design of complex neural network
		structures promises to improve performance of deep learning
		architectures beyond human ability to do so.
		
	\end{abstract}

	\maketitle
	
	\begin{figure*}
		\includegraphics[height=1.8in, width=7in]{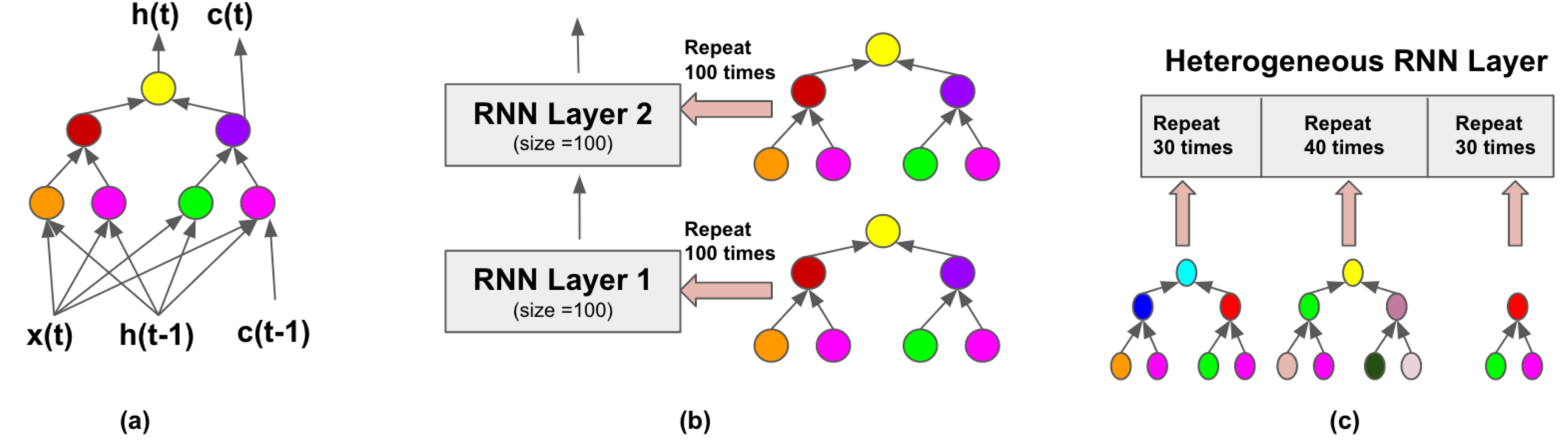}
		\caption{(a)Tree based representation of the recurrent node. Tree outputs $h(t)$ and $c(t)$ are fed as inputs in the next time step. (b) In standard recurrent network, the tree node is repeated several times to create each layer in a multi-layered network. Different node colors depict various element activations. (c) The heterogeneous recurrent layer consists of different types of recurrent nodes.}
		\label{fg:node2network}
	\end{figure*}
	\section{Introduction}
	
	In many areas of engineering design, the systems have become so
	complex that humans can no longer optimize them, and instead,
	automated methods are needed. This has been true in VLSI design for a
	long time, but it has also become compelling in software engineering:
	The idea in "programming by optimization" is that humans should design
	only the framework and the details should be left for automated
	methods such as optimization \cite{hoos:acm12}.  Recently similar
	limitations have started to emerge in deep learning. The neural
	network architectures have grown so complex that humans can no longer
	optimize them; hyperparameters and even entire architectures are now
	optimized automatically through gradient descent \cite{marcin:nips}, Bayesian parameter
	optimization \cite{garnett:nips}, reinforcement learning \cite{zoph} \cite{baker:RL}, and evolutionary computation \cite{risto} \cite{esteban}, \cite{chrisantha}.  Improvements from such
	automated methods are significant: the structure of the network
	matters.
	
	This paper shows that the same approach can be used to improve
	architectures that have been used essentially unchanged for decades. The
	case in point is the Long Short-Term Memory (LSTM) network
	\cite{lstm}. It was originally proposed in 1992;
	with the vastly increased computational power, it has recently been
	shown a powerful approach for sequential tasks such as speech
	recognition, language understanding, language generation, and machine
	translation, in some cases improving performance 40\% over traditional
	methods \cite{bahdanau:iclr}. The basic LSTM structure has changed very little in
	this process, and thorough comparisons of variants concluded that
	there's little to be gained by modifying it further \cite{klaus:arxiv14} \cite{jozefowicz:icml}.
	
	However, very recent studies on metalearning methods such as neural
	architecture search and evolutionary optimization have shown that LSTM
	performance can be improved by complexifying it further
	\cite{zoph} \cite{risto}. This paper develops a new method along
	these lines, recognizing that a large search space where significantly
	more complex node structures can be constructed could be
	beneficial. The method is based on a tree encoding of the node
	structure so that it can be efficiently searched using genetic
	programming. Indeed, the approach discovers significantly more complex
	structures than before, and they indeed perform significantly better:
	Performance in the standard language modeling benchmark, where the
	goal is to predict the next word in a large language corpus, is
	improved by 6 perplexity points over the standard LSTM \cite{zaremba:arxiv14}, and
	0.9 perplexity points over the previous state of the art,
	i.e.\ reinforcement-learning based neural architecture search
	\cite{zoph}.
	
	These improvements are obtained by constructing a homogeneous layered
	network architecture from a single gated recurrent node design. A
	second innovation in this paper shows that further improvement can be
	obtained by constructing such networks from multiple different
	designs.  As a first step, allocation of different kinds of LSTM nodes
	into slots in the network is shown to improve performance by another 0.5
	perplexity points. This result suggests that further improvements are
	possible with more extensive network-level search.
	
	A third contribution of this paper is to show that evolution of neural
	network architectures in general can be speeded up significantly by
	using an LSTM network to predict the performance of candidate neural
	networks. After training the candidate for a few epochs, such a
	Meta-LSTM network predicts what performance a fully trained network
	would have. That prediction can then be used as fitness for the
	candidate, speeding up evolution fourfold in these experiments. A
	fourth contribution is to encourage exploration by using an archive of
	already-explored areas. The effect is similar to that of novelty
	search, but does not require a separate novelty objective, simplifying
	the search.
	
	Thus, evolutionary optimization of complex deep learning architectures
	is a promising approach that can yield significant improvements beyond
	human ability to do so.
	
	\section{Background and Related Work}
	
	In recent years, LSTM-based recurrent networks have been used to achieve
	strong results in the supervised sequence learning problems
	such as in speech recognition [10] and machine translation \cite{bahdanau:iclr}. 
	Further
	techniques have been developed to improve performance of these models
	through ensembling \cite{zaremba:arxiv14}, 
	shared embeddings \cite{zilly:highwaylstm} and dropouts \cite{rnndropout}. 
	
	In contrast, previous studies have shown that
	modifying the LSTM design itself did not provide any significant performance gains \cite{bayer:icann} \cite{cho:gru} \cite{jozefowicz:icml}.
	However, a recent paper from Zoph et. al. \cite{zoph} showed that policy 
	gradients can be used to train a LSTM network to find better LSTM designs. The network is rewarded based
	on the performance of the designs it generates. While this approach can be used to create new
	designs that perform well, its exploration ability is limited (as
	described in more detail in Section 3.3). 
	
	Neuroevolution methods like NEAT \cite{stanley:ec02} are an alternative to policy gradient approaches, and have also been shown to be sucessful in the architecture search problem \cite{risto} \cite{esteban}. 
	For instance, Cartesian genetic programming was recently used to achieve state of the art results in CIFAR-10 \cite{cgp}. 
	Along similar lines, a tree based variant of genetic programming is used in this paper to evolve recurrent nodes. These trees can grow in structure and can be pruned as well, thus providing a flexible representation. 
	
	Novelty search is a particularly useful technique to increase
	exploration in evolutionary optimization
	\cite{joel:novelty}.  Novelty is often cast as a secondary
	objective to be optimized.  It allows searching in areas that do not
	yield immediate benefit in terms of fitness, but make it possible to
	discover stepping stones that can be combined to form better solutions
	later. This paper proposes an alternative approach: keeping an archive
	of areas already visited and exploited, achieving similar goals
	without additional objectives to optimize.

	Most architecture search methods reduce compute time by evaluating individuals only after partial training \cite{cgp} \cite{esteban}. This paper proposes a meta LSTM framework to predict final network performance based on partial training results. 
	
	These techniques are described in detail in the next section.

	\section{Methods}
	Evolving recurrent neural networks is an interesting problem
	because it requires searching the architecture of both the node
	and the network. As shown by recent research \cite{zoph} \cite{zilly:highwaylstm}, the
	recurrent node in itself can be considered a deep network. In this paper, Genetic Programming (GP) is
	used to evolve such node architectures.  In the first experiment,
	the overall network architecture is fixed i.e.\ constructed by
	repeating a single evolved node to form a layer (Figure\ref{fg:node2network}(b)). In the second,
	it is evolved by combining several different types of nodes into
	a layer (Figure\ref{fg:node2network}(c)). In the future more
	complex coevolution approaches are also possible.
	
	Evaluating the evolved node and network is costly. Training the
	network for 40 epochs takes two hours on a 1080 NVIDIA GPU. A
	sequence to sequence model called meta-LSTM is developed to speed
	up evaluation. Following sections describe these methods in
	detail.
	
	\begin{figure*}
		\includegraphics[height=1.9in, width=6in]{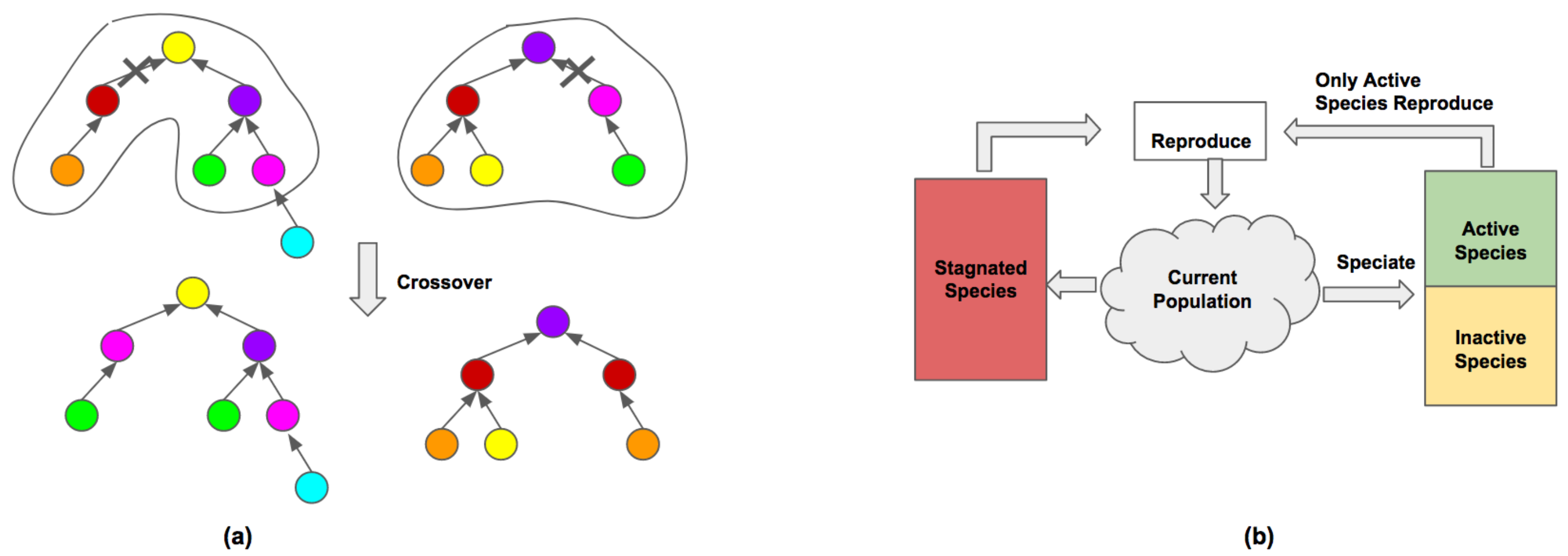}
		\caption{(a) Homologous crossover in GP - the two trees on the top look different but in-fact they are almost mirror images of each other. These two trees will therefore belong in the same species. The line drawn around the trees marks the homologous regions between the two. A crossover point is randomly selected and one point crossover is performed. The bottom two networks are the resultant offsprings.  (b) An archive of stagnant species called Hall of Shame (shown in red) is built during evolution. This archive is looked up during reproduction, to make sure that newly formed offsprings do not belong to any of the stagnant species. At a time, only 10 species are actively evaluated (shown in green). This constraint ensures that active species get enough spawns to ensure a comprehensive search in its vicinity before it is added to the Hall of Shame. Offsprings that belong to new species are pushed into a inactive species list (shown in yellow) and are only moved to the active list whenever an active species moves to Hall of Shame.}
		\label{fg:crossover}
	\end{figure*}
	
	\subsection{Genetic Programming for Recurrent Nodes}
	
	As shown in Figure\ref{fg:node2network}(a), a recurrent node can be represented as a
	tree structure, and GP can therefore be used to evolve it.
	However, standard GP may not be sufficiently powerful to do
	it. In particular, it does not maintain sufficient diversity in
	the population.  Similar to the GP-NEAT approach by Trujillo et al.
	\cite{neatgp}, it can be augmented with ideas from NEAT speciation.

	A recurrent node usually has two types of outputs. The first, denoted
	by symbol $h$ in Figure\ref{fg:node2network} (a), is the main recurrent output. The second,
	often denoted by $c$, is the native memory cell output. The $h$ value
	is weighted and fed to three locations: (1) to the higher layer of the
	network at the same time step, (2) to other nodes in the network at
	the next time step, and (3) to the node itself at the next time step.
	Before propagation, $h$ are combined with weighted activations from
	the previous layer, such as input word embeddings in language
	modeling, to generate eight node inputs (termed as base eight by Zoph
	et al. \cite{zoph}). In comparison, the standard LSTM node has four
	inputs (see Figure\ref{fg:solution}(a)). The native memory cell output is fed back, without weighting,
	only to the node itself at the next time step. The connections within a recurrent cell are not trainable by backpropagation and they all carry a fixed weight of 1.0. 
	
	Thus, even without an explicit recurrent loop, the recurrent node can
	be represented as a tree. There are two type of elements in the tree:
	(1) linear activations with arity two (add, multiply), and (2)
	non-linear activations with arity one (tanh, sigmoid, relu).
	
	There are three kind of mutation operations in the experiments: (1)
	Mutation to randomly replace an element with an element of the same
	type, (2) Mutation to randomly inserts a new branch at a random
	position in the tree. The subtree at the chosen position is used as
	child node of the newly created subtree. (3) Mutation to shrink the
	tree by choosing a branch randomly and replacing it with one of the
	branch's arguments (also randomly chosen).
	
	One limitation of standard tree is that it can have only a single
	output: the root. This problem can be overcome by using a modified
	representation of a tree that consists of Modi outputs \cite{modi}. In this
	approach, with some probability $p$ (termed modirate), non-root nodes
	can be connected to any of the possible outputs. A higher modi rate
	would lead to many sub-tree nodes connected to different outputs.
	A node is assigned modi (i.e. connected to memory cell outputs c or d) only if its sub-tree has a path from native memory cell inputs.  
	
	This representation allows searching for a wide range of recurrent
	node structures with GP.

	\subsection{Speciation and Crossover}
	
	One-point crossover is the most common type of crossover in GP.
	However, since it does not take into account the tree structure, it
	can often be destructive. An alternative approach, called homologous
	crossover \cite{homogp}, is designed to avoid this problem by crossing over the
	common regions in the tree.  Similar tree structures in the population
	can be grouped into species, as is often done in NEAT \cite{neatgp}.  Speciation achieves two objectives: (1) it makes homologous
	crossover effective, since individuals within species are similar, and
	(2) it helps keep the population diverse, since selection is carried
	out separately in each species.  A tree distance metric proposed by Tujillo et al. (\cite{neatgp}) is used to
	determine how similar the trees are:
	
	\begin{equation}
	\delta(T_i, T_j) = \beta \frac{N_{i,j} - 2n_{S_i,j}}{N_{i,j} - 2} + (1-\beta)  \frac{D_{i,j} - 2d_{S_i,j}}{D_{i,j} - 2} ,
	\label{eq:treedist}
	\end{equation}
	where:
	
	$n_{T_x}$ = number of nodes in GP tree $T_x,$ 
	
	$d_{T_x}$ = depth of GP tree $T_x,$
	
	$S_{i,j}$ = shared tree between $T_i$ and $T_j,$
	
	$N_{i,j}$ = $n_{T_i} + n_{T_j},$
	
	$D_{i,j}$ = $d_{T_i} + d_{T_j},$
	
	$\beta \in [0,1],$
	
	$\delta \in [0,1].$ \\
	
	On the right-hand side of Equation~\ref{eq:treedist}, the first term measures the difference with respect to size, while the second term measures the difference in depth. Thus, setting $\beta = 0.5$ gives an equal importance to size and depth. Two trees will have a distance of zero if their structure is the
	same (irrespective of the actual element types).  
	
    In most GP
	implementations, there is a concept of the left and the right
	branch. A key extension in this paper is that the tree distance is
	computed by comparing trees after all possible tree rotations, i.e.\
	swaps of the left and the right branch.  Without
	such a comprehensive tree analysis, two trees that are mirror images
	of each other might end up into different species.  This approach
	reduces the search space by not searching for redundant trees. It also
	ensures that crossover can be truly homologous Figure\ref{fg:crossover} (a).
	
	The structural mutations in GP, i.e.\ insert and shrink, can lead to
	recycling of the same strcuture across multiple generations. In order
	to avoid such repetitions, an archive called Hall of Shame is
	maintained during evolution (Figure\ref{fg:crossover}(b)). This archive consists of individuals
	representative of stagnated species, i.e.\ regions in the architecture
	space that have already been discovered by evolution but are no longer
	actively searched. During reproduction, new offsprings are repeatedly
	mutated until they result in an individual that does not belong to
	Hall of Shame. Mutations that lead to Hall of Shame are not discarded,
	but instead used as stepping stones to generate better individuals.
	Such memory based evolution is similar to novelty search. However,
	unlike novelty search \cite{joel:novelty}, there is no additional fitness objective,
	simply an archive.

	\subsection{Search Space: Node}
	GP evolution of recurrent nodes starts with a simple fully connected
	tree. During the course of evolution, the tree size increases due to
	insert mutations and decreases due to shrink mutations. The maximum
	possible height of the tree is fixed at 15. However, there is no
	restriction on the maximum width of the tree.
	
	The search space for the nodes is more varied and several orders of
	magnitude larger than in previous approaches.  More specifically, the
	main differences from the state-of-the-art Neural Architecture Search
	(NAS) \cite{zoph} are: (1) NAS searches for trees of fixed height 10
	layers deep; GP searches for trees with height varying between six
	(the size of fully connected simple tree) and 15 (a constraint added
	to GP). (2) Unlike in NAS, different leaf elements can occur at
	varying depths in GP. (3) NAS adds several constraint to the tree
	structure. For example, a linear element in the tree is always
	followed by a non-linear element. GP prevents only consecutive
	non-linearities (they would cause loss of information since the
	connections within a cell are not weighted). (4) In NAS, inputs to the
	tree are used only once; in GP, the inputs can be used multiple times
	within a node.
	
	Most gated recurrent node architectures consist of a single native
	memory cell (denoted by output $c$ in Figure\ref{fg:node2network}(a)). This memory cell
	is the main reason why LSTMs perform better than simple RNNs. One key
	innovation introduced in this paper is to allow multiple native memory
	cells within a node. The memory cell output is fed back as input in
	the next time step without any modification, i.e.\ this recurrent loop
	is essentially a skip connection. Adding another memory cell in the
	node therefore does not effect the number of trainable parameters: It
	only adds to the representational power of the node.

	\subsection{Search Space: Network}
	Standard recurrent networks consist of layers formed by repetition of
	a single type of node.  However, the search for better recurrent nodes
	through evolution often results in solutions with similar task
	performance but very different structure. Forming a recurrent layer by
	combining such diverse node solutions is potentially a powerful idea,
	related to the idea of ensembling, where different models are combined
	together to solve a task better.
	
	In this paper, such heterogenous recurrent networks are constructed by
	combining diverse evolved nodes into a layer (Figure\ref{fg:node2network}(c)). A candidate population
	is created that consists of top-performing evolved nodes that are
	structurally very different from other nodes. The structure difference
	is calculated using the tree distance formula detailed previously. Each
	heterogenous layer is constructed by selecting nodes randomly from the
	candidate population. Each node is repeated 20 times in a layer; thus,
	if the layer size is e.g.\ 100, it can consist of five different node
	types, each of cardinality 20.
	
	The random search is an initial test of this idea. As described in
	Section 5, in the future the idea is to search for
	such heterogenous recurrent networks using a genetic algorithm as
	well.

	\begin{figure}
		\includegraphics[height=2in, width=3in]{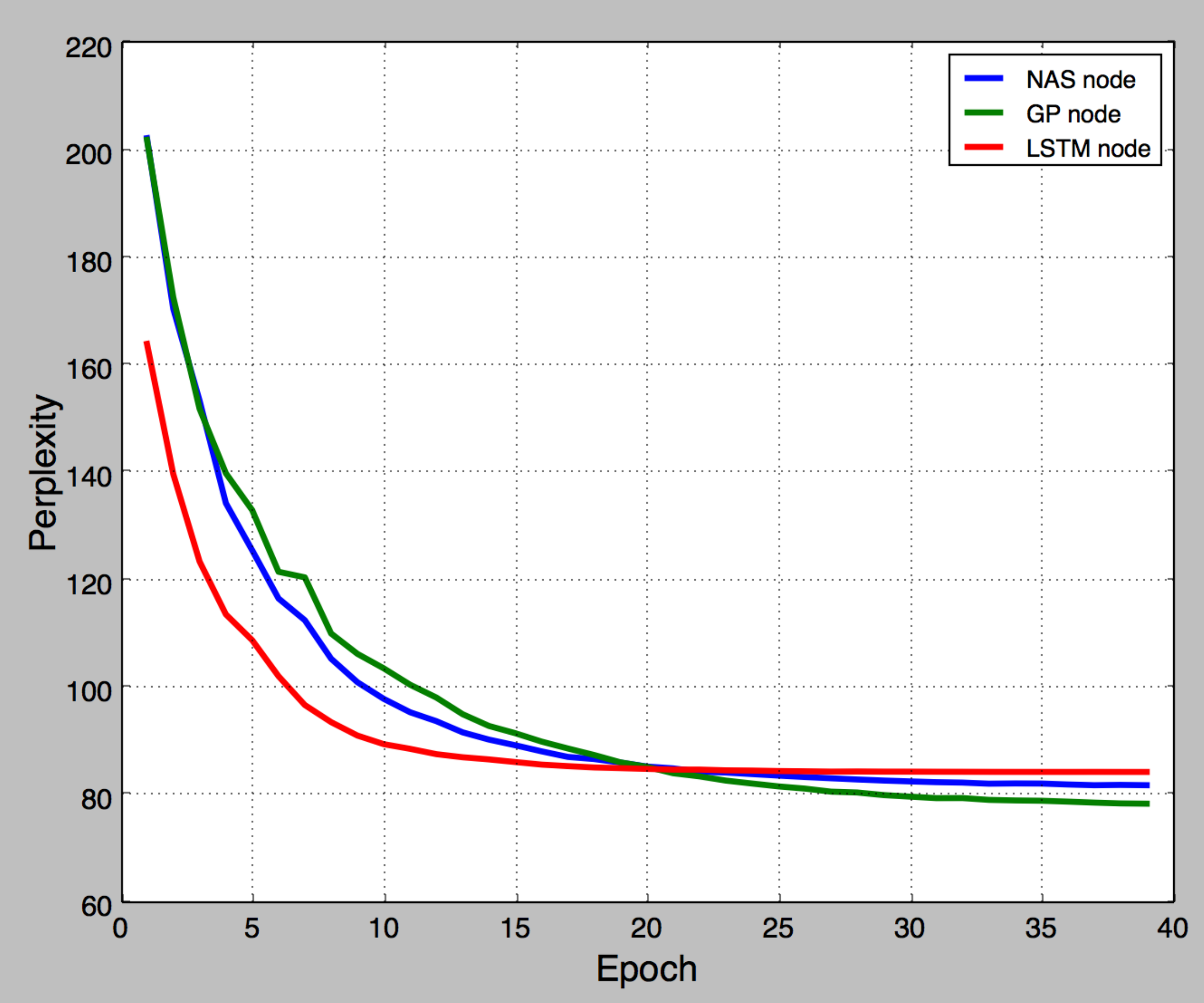}
		\caption{Learning curve comparison of LSTM node, NAS node and GP nodes. Y-axis is the validation perplexity (lower is better) and X-axis is the epoch number. Notice that LSTM node learns quicker than the other two initially but eventually settles at a larger perplexity value. This graph demonstrates that the strategy to determine network fitness using partial training (say based on epoch 10 validation perplexity) is faulty. A fitness predictor model like meta LSTM can overcome this problem.}
		\label{fg:loss_curve}
	\end{figure}
	
	\subsection{Meta-LSTM for Fitness Prediction}
	
	In both node and network architecture search, it takes about two hours
	to fully train a network until 40 epochs. With sufficient computing
	power it is possible to do it: for instance Zoph et al. \cite{zoph} used
	800 GPUs for training multiple such solutions in parallel. However,
	if training time could be shortened, no matter what resources are
	available, those resources could be used better.
	
	A common strategy for such situations is early stopping \cite{cgp}, i.e.\ selecting networks based on partial training. For
	example in case of recurrent networks, the training time would be cut
	down to one fourth if the best network could be picked based on the
	10th epoch validation loss instead of 40th. Figure\ref{fg:loss_curve} demonstrates that
	this is not a good strategy, however. Networks that train faster in
	the initial epochs often end up with a higher final loss.
	
\begin{figure*}
		\includegraphics[height=1.9in, width=6in]{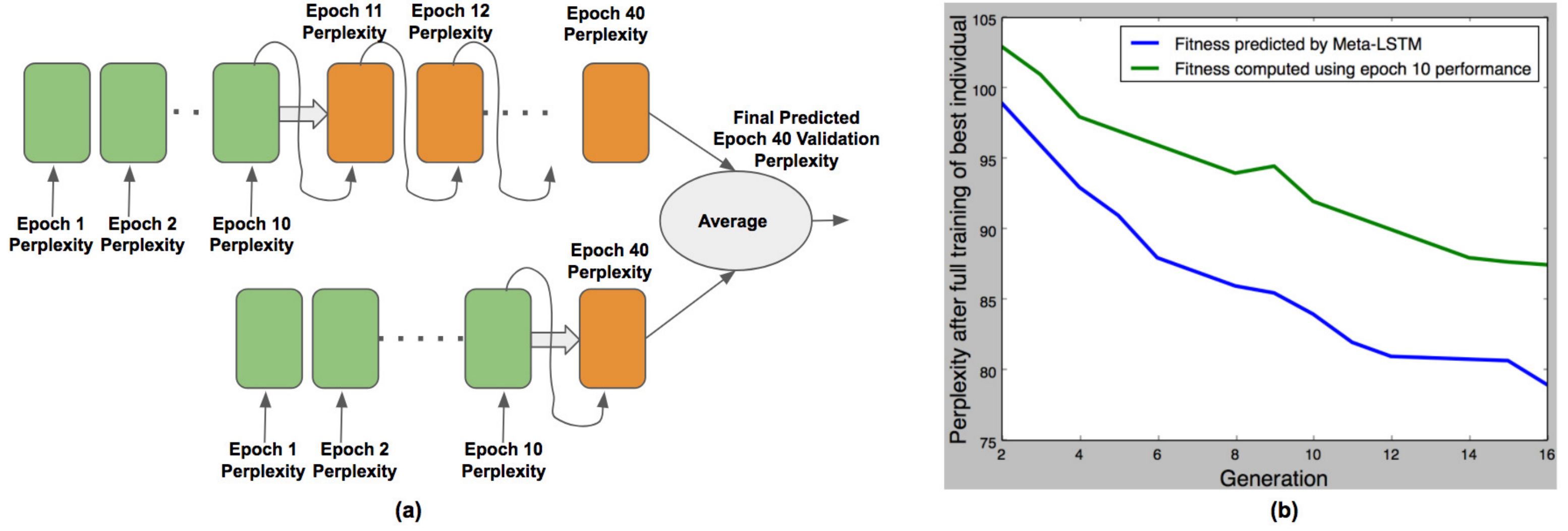}
		\caption{(a) Meta LSTM model: this is a sequence to sequence (seq2seq) model that takes the validation perplexity of the first 10 epochs as sequential input and predicts the validation perplexity at epoch 40. The green rectangles denote the encoder and the orange rectangles denote the decoder. Two variants of the model are averaged to generate one final prediction. In one variant (top), the decoder length is 30 and in the other variant (bottom), the decoder length is 1.  (b) Meta LSTM performance: Two evolution experiments are conducted - one, where epoch 10 validation perplexity of the network is used as the fitness and second, where the value predicted by meta LSTM is used as the network fitness. After evolution has completed, the best individuals from each generation are picked and fully trained till epoch 40. For both the experiments, this graph plots the epoch 40 performance of the best network in a given generation. The plot shows that as evolution progresses, meta LSTM framework selects better individuals. }
		\label{fg:metalstm}
	\end{figure*}

	To overcome costly evaluation and to speed up evolution, a Meta-LSTM
	framework for fitness prediction was developed.  Meta-LSTM is a
	sequence to sequence model \cite{seq2seq} that consists of an encoder RNN and
	a decoder RNN (see Figure\ref{fg:metalstm}(a)). Validation perplexity of the first 10 epochs is
	provided as sequential input to the encoder, and the decoder is
	trained to predict the validation loss at epoch 40 (show figure).
	Training data for these models is generated by fully training sample
	networks (i.e.\ until 40 epochs).  The loss is the mean absolute error
	percentage at epoch 40. This error measure is used instead of mean
	squared error because it is unaffected by the magnitude of perplexity
	(poor networks can have very large perplexity values that overwhelm
	MSE). The hyperparameter values of the Meta-LSTM were selected based
	on its performance in the validation dataset. The best configuration
	that achieved an error rate of 3\% includes an ensemble of two seq2seq
	models: one with a decoder length of 30 and the other with a decoder
	length of 1 (figure).

	Recent approaches to network performance prediction include Bayesian modeling (\cite{klein:iclr2017})
	and regression curve fitting \cite{fitness_pred:iclr2018}.  The learning curves for which 
	the above methods are deployed are much simpler as compared to the learning curves of structures discovered by evolution (see Appendix). 
	Note that Meta-LSTM is trained separately and only deployed for use
	during evolution. Thus, networks can be partially trained with a
	4$\times$ speedup, and assessed with near-equal accuracy as with full
	training.
	
	\section{Experiments}
	
	Neural architectures were constructed for the language modeling task,
	using Meta-LSTM as the predictor of training performance. In the first
	experiment, homogeneous networks were constructed from single evolved
	nodes, and in the second, heterogeneous networks that consisted of
	multiple evolved nodes.
	
	\subsection{Natural Language Modeling Task}
	
	Experiments focused on the task of predicting the next word in the
	Penn Tree Bank corpus (PTB), a well-known benchmark for language
	modeling \cite{marcus:ptb}. LSTM architectures in general tend
	to do well in this task, and improving them is difficult \cite{zaremba:arxiv14} \cite{jozefowicz:icml} \cite{rnndropout}. The dataset consists
	of 929k training words, 73k validation words, and 82k test words, with
	a vocabulary of 10k words. During training, successive minibatches of
	size 20 are used to traverse the training set sequentially.
	
	\subsection{Music Modeling Task}
	Music consists of a sequence of notes that often exhibit temporal dependence. Predicting future notes based 
	on the previous notes can therefore be treated as a sequence prediction problem. 
	Similar to natural language, musical structure can be captured using a music language model (MLM).
	Just like natural language models form an important component of speech recognition systems, 
	polyphonic music language model are an integral part of Automatic music transcription (AMT). 
	AMT is defined as
	the problem of extracting a symbolic representation from
	music signals, usually in the form of a time-pitch representation
	called piano-roll, or in a MIDI-like representation.

	MLM predict the probability distribution of the notes in the next time step. Multiple notes can be turned-on
	at a given time step for playing chords.
	The input is a piano-roll representation, in the form of an
	88$XT$ matrix $M$, where $T$ is the number of timesteps, and
	88 corresponds to the number of keys on a piano, between
	MIDI notes A0 and C8. M is binary, such that $M[p, t] = 1$
	if and only if the pitch $p$ is active at the timestep $t$. In
	particular, held notes and repeated notes are not differentiated.
	The output is of the same form, except it only has
	$T-1$ timesteps (the first timestep cannot be predicted since
	there is no previous information). 
	
	Piano-midi.de dataset is used as the benchmark data. This dataset currently holds 307 pieces of classical
	piano music from various composers. It was made by manually
	editing the velocities and the tempo curve of quantised
	MIDI files in order to give them a natural interpretation
	and feeling (~\cite{music:ismir2017}).
	MIDI files encode explicit timing, pitch,
	velocity and instrumental information of the musical score. 
	\subsection{Network Training Details}
	
	During evolution, each network has two layers of 540 units each, and
	is unrolled for 35 steps. The hidden states are initialized to zero;
	the final hidden states of the current minibatch are used as the
	initial hidden states of the subsequent minibatch. The dropout
	rate is 0.4 for feedforward connections and 0.15 for recurrent
	connections \cite{rnndropout}. The network weights have L2 penalty of 0.0001. The
	evolved networks are trained for 10 epochs with a learning rate of 1;
	after six epochs the learning rate is decreased by a factor of 0.9
	after each epoch. The norm of the gradients (normalized by minibatch
	size) is clipped at 10. Training a network for 10 epochs takes about
	30 minutes on an NVIDIA 1080 GPU. The following experiments were
	conducted on 40 such GPUs.
	
	The Meta-LSTM consists of two layers, 40 nodes each. To generate
	training data for it, 1000 samples from a preliminary node evolution
	experiment was obtained, representing a sampling of designs that
	evolution discovers. Each of these sample networks was trained for 40
	epochs with the language modeling training set; the perplexity on the
	language modeling validation set was measured in the first 10 epochs,
	and at 40 epochs. The Meta-LSTM network was then trained to predict
	the perplexity at 40 epochs, given a sequence of perplexity during the
	first 10 epochs as input. A validation set of 500 further networks was
	used to decide when to stop training the Meta-LSTM, and its accuracy
	measured with another 500 networks.
	
	In line with Meta-LSTM training, during evolution each candidate is
	trained for 10 epochs, and tested on the validation set at each
	epoch. The sequence of such validation perplexity values is fed into
	the trained meta-LSTM model to obtain its predicted perplexity at
	epoch 40; this prediction is then used as the fitness for that
	candidate. The individual with the best fitness after 30 generations is
	scaled to a larger network consisting of 740 nodes in each layer. This
	setting matches the 32 Million parameter configuration used by Zoph et
	al. \cite{zoph}. A grid search over drop-out rates is carried out to
	fine-tune the model. Its performance after 180 epochs of training is
	reported as the final result (Table~\ref{tab:results})
	
	\begin{figure*}
		\includegraphics[height=2.7in, width=6.8in]{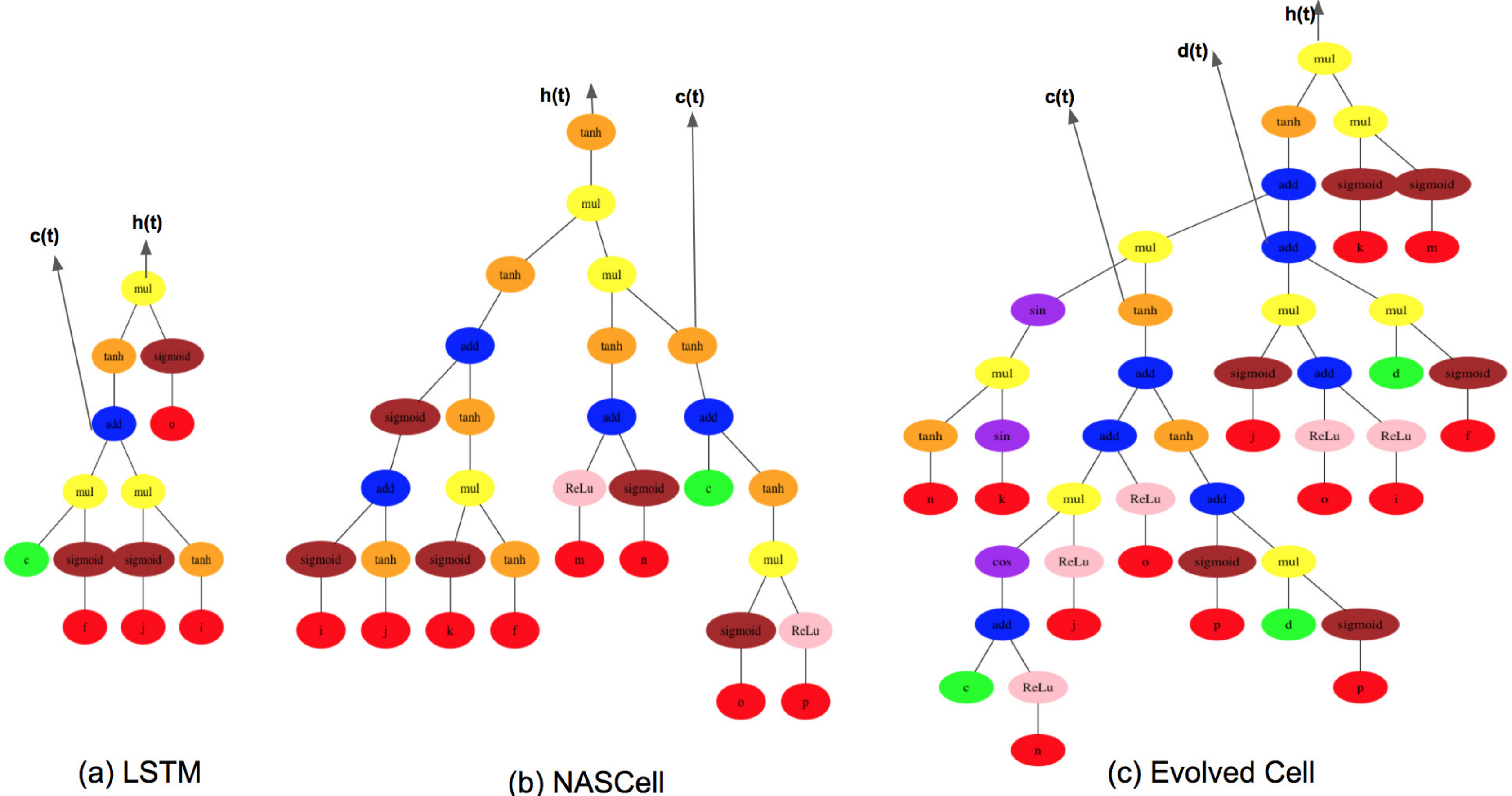}
		\caption{(a) Comparing Evolved recurrent node with NASCell and LSTM. The green input elements denote the native memory cell outputs from the previous time step ($c$, $d$). The red colored inputs are formed after combining the node output from the previous time step $h(t-1)$ and the new input from the current time step $x(t)$. In all three solutions, the memory cell paths include relatively few non-linearities. The evolved node utilizes the extra memory cell in different parts of the node. GP evolution also reuses inputs unlike the NAS and LSTM solution. Evolved node also discovered LSTM like output gating.}
		
		\label{fg:solution}
	\end{figure*}
	\subsection{Experiment 1: Evolution of Recurrent Nodes}
	
	A population of size 100 was evolved for 30 generations with a
	crossover rate of 0.6, insert and shrink mutation probability of 0.6
	and 0.3, respectively, and modi rate (i.e.\ the probability that a
	newly added node is connected to memory cell output) of 0.3. A
	compatibility threshold of 0.3 was used for speciation; species is
	marked stagnated and added to the Hall of Shame if the best fitness
	among its candidates does not improve in four generations. Each node
	is allowed to have three outputs: one main recurrent output ($h$) and
	two native memory cell outputs ($c$ and $d$).

	The best evolved node is shown Figure\ref{fg:solution}.  The evolved node reuses inputs as well as utilize the extra memory cell pathways. As shown in Table~\ref{tab:results}, the evolved node (called GP Node evolution in the table) achieves a test performance of 68.2 for 20 Million parameter configuration on Penn Tree Bank. This is 2.8 perplexity points better than the test performance of the node discovered by NAS (Zoph(2016) in the table) in the same configuration. Evolved node also outperforms NAS in the 32 Million configuration (68.1 v/s. 66.5). Recent work has shown that sharing input and output embedding weight matrices of neural network language models improves performance \cite{ofir:arxiv16}. The experimental results obtained after including this method are marked as shared embeddings in Table~\ref{tab:results}. 
	
	It is also important to understand the impact of using meta LSTM in evolution. For this purpose, an additional evolution experiment was conducted, where each individual was assigned a fitness equal to its 10th epoch validation perplexity. As evolution progressed, in each generation, the best individual was trained fully till epoch 40. Similarly, the best individual from a evolution experiment with meta LSTM enabled was fully trained. The epoch 40 validation perplexity in these two cases has been plotted in Figure\ref{fg:metalstm}(b). This figure demonstrates that individuals that are selected based upon meta LSTM prediction perform better than the ones selected using only partial training.

	\begin{table}
		\caption{Single Model Perplexity on Test set of Penn Tree Bank. Node evolved using GP outperforms the node discovered by NAS (Zoph(2016)) and Recurrent Highway Network (Zilly et al. (2016)) in various configurations.}
		\label{tab:results}
		\begin{tabular}{ccl}
			\toprule
			Model&Parameters&Test Perplexity\\
			\midrule
			Gal (2015) - Variational LSTM & 66M & 73.4\\
			Zoph (2016)  & 20M & 71.0\\ 
			GP Node Evolution  & 20M & 68.2\\
			Zoph (2016) & 32M & 68.1\\
			GP Node Evolution  & 32M & 66.5\\
			Zilly et al. (2016) , shared embeddings & 24M & 66.0\\
			Zoph (2016), shared embeddings & 25M & 64.0\\
			GP Evolution, shared embeddings & 25M & 63.0\\
			Heterogeneous, shared embeddings & 25M & 62.2\\
			Zoph (2016), shared embeddings & 54M & 62.9\\
			\bottomrule
		\end{tabular}
	\end{table}

	\subsection{Experiment 2: Heterogeneous Recurrent Networks}
	
	Top 10\% of the population from 10 runs of Experiment 1 was collected
	into a pool 100 nodes. Out of these, 20 that were the most diverse,
	i.e.\ had the largest tree distance from the others, were selected for
	constructing heterogeneous layers (as shown in Figure\ref{fg:node2network}(c)).  Nodes were chosen from this pool
	randomly to form 2000 such networks. Meta-LSTM was again used to speed
	up evaluation.
	
	After hyperparameter tuning, the best network (for 25 Million parameter configuration )achieved a perplexity of
	62.2, i.e.\ 0.8 better than the homogeneous network constructed from
	the best evolved node. This network is also 0.7 perplexity point better than the best NAS network double its size (54 Million parameters). Interestingly, best heterogeneous network was
	also found to be more robust to hyperparameter changes than the
	homogeneous network. This result suggests that diversity not only
	improves performance, but also adds flexibility to the internal
	representations. The heterogeneous network approach therefore forms a
	promising foundation for future work, as discussed next.
	
	\subsection{Experiment 3: Music Modeling}
	Piano-midi.de dataset is used as a benchmark data. 
	This data is divided into train (60\%), test (20\%) and validation (20\%) sets. 
	The music language model consists of a single recurrent layer of width 128.
	The input and output layers are 88 wide each. 
	The network is trained for 50 epochs with Adam at a learning rate of 0.01. 
	The network is trained by minimizing cross entropy between the output of the network and the ground truth. For evaluation, F1 score is computed on the test data. F1 score is the harmonic mean of precision and recall (higher is better). Since the network is smaller, regularization is not required. 
	
	Note, this setup is similar to the one presented in ~\cite{music:ismir2017}. The goal of this experiment is not to achieve state-of-the-art results but instead is to perform apples-to-apples comparison between LSTM node and evolved node (discovered for language) in a new domain i.e. music. 
	
	Three networks were constructed: first with LSTM nodes, second NAS node and the third with evolved node. All 
	the three networks were trained under the same setting as described in the previous section. The F1 score of each of the three models is 
	shown in Table ~\ref{tab:music_results}. LSTM node outperforms both NAS and evolved nodes. This result is interesting because both NAS and evolved nodes significantly outperformed LSTM node in the language-modeling task. This results suggests that NAS and evolved nodes are custom solution for a specific domain. 
	
	\begin{table}
		\centering
		\caption{F1 scores computed on Piano-Midi dataset. LSTM outperforms both the evolved node and NAS node.}
		\label{tab:music_results}
		\begin{tabular}{ccl}
			\toprule
			Model&F1 score\\
			\midrule
			LSTM & 0.548\\
			Zoph (2016)  & 0.48\\ 
			GP Evolution (Language) & 0.49\\
			GP Evolution (Music) & 0.599\\
			\bottomrule
		\end{tabular}
	\end{table}
	
	The framework developed for evolving recurrent nodes for natural language can be transferred to other domains like music. The setup could be very similar i.e. at each generation a population of recurrent nodes represented as trees will be evaluated for their performance in the music domain. The validation performance of the network constructed from the respective tree node will be used as the node fitness. The performance measure of the network in music domain is the F1 score, therefore, it is used as the network fitness value. 
	
\begin{figure}
		\includegraphics[height=2.7in, width=3.3in]{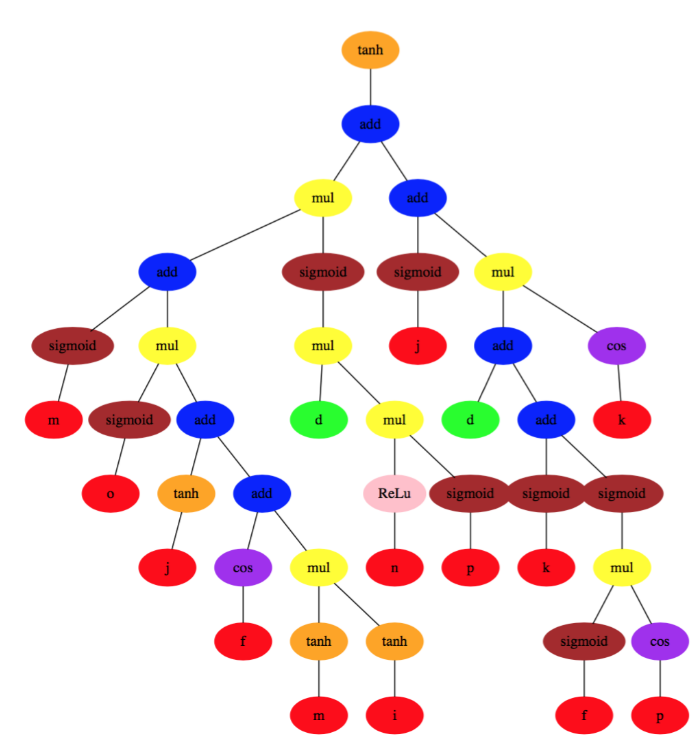}
		\caption{Evolved Node for Music. The node evolved to solve the music task is very different from the node for the natural language task. For example, this node only uses a single memory cell (green input $d$ in the figure) unlike the language node that used both $c$ and $d$. This results indicates that 'architecture does matter' and that custom evolved solution perform better than hand-designed ones.}
		\label{fg:music_solution}
	\end{figure}

	The evolution parameters are the same as that used for natural language modeling. Meta-lstm is not used for this evolution experiment because the run-time of each network is relatively small (< 600 seconds). Result from evolving custom node for music is shown in Table ~\ref{tab:music_results}. The custom node (GP Evolution (Music)) achieves an improvement of five points in F1 score over LSTM (see Figure~\ref{fg:music_solution}). 
	
	\section{Discussion and Future Work}
	
	The experiments in this paper demonstrate how evolutionary
	optimization can discover improvements to designs that have been
	essentially unchanged for 25 years.  Because it is a population-based
	method, it can harness more extensive exploration than other
	meta-learning techniques such as reinforcement learning, Bayesian
	parameter optimization, and gradient descent. It is therefore in a
	position to discover novel, innovative solutions that are difficult to
	develop by hand or through gradual improvement.
	
	Population in evolutionary method can also be used to amplify another
	technique that has become common in deep learning in general and LSTM
	network systems in particular: ensembling \cite{zaremba:arxiv14}.  The diverse
	solutions in the population are a natural source if individuals for
	the ensemble: they are good but different, and their collective
	behavior may be used to construct behavior that's better than any
	individual alone. The members of the current population may already be
	used to this effect, however, the approach can be further improved in
	two ways: First, diversity can be explicitly encouraged, for instance
	by making it a secondary objective, resulting in more exploration and
	a more varied base for the ensemble. Second, the population can be
	evolved to maximize ensemble fitness directly. That is, ensembles are
	formed during evolution, and their performance is shared as the
	fitness of the individuals in the ensemble. In this manner, not just
	diversity, but useful diversity is explicitly rewarded, which should
	lead to better ensembles.
	
	The current experiments focused on optimizing the structure of the
	gated recurrent nodes, cloning them into a fixed layered architecture
	to form the actual network. The simple approach of forming
	heterogeneous layers by choosing from a set of different nodes was
	shown to improve the networks further. A compelling next step is thus
	to evolve the network architecture as well, and further, coevolve it
	together with the LSTM nodes. A similar approach has proven powerful
	in designing large deep learning networks composed of component
	networks (the CoDeepNEAT method \cite{risto}), and should apply to the
	level of networks and nodes as well.  The search space can be then
	expanded further by allowing evolution to discover various skip
	connections between nodes at the same and different layers. These
	extensions follow naturally from current work in deep neuroevolution,
	and should increase the performance of deep learning methods further
	beyond human design.

	\section{Conclusions}
	Evolutionary optimization of LSTM nodes can be used to discover new
	variants that perform significantly better than the original 25-year
	old design. The tree-based encoding and genetic programming approach
	makes it possible to explore larger design spaces efficiently,
	resulting in structures that are more complex and more powerful than
	those discovered by hand or through reinforcement-learning based
	neural architecture search. The approach can be further enhanced by
	optimizing the network level as well, in addition to the node
	structure, by training an LSTM network to estimate the final
	performance of candidates instead of having to train them
	fully, and by encouraging novelty through an archive. Evolutionary
	neural architecture search is therefore a promising approach to
	extending the abilities of deep learning networks to ever more
	challenging tasks.
	%
	%

	
	\bibliographystyle{ACM-Reference-Format}
	\bibliography{lstm} 
	
\end{document}